\documentclass[letterpaper]{article}
\usepackage{times}
\usepackage{latexsym}

\usepackage{url}  
\usepackage{graphicx}  

\usepackage{amsfonts}       

\usepackage{amsmath}
\usepackage{amssymb}
\usepackage{bbm}
\usepackage{bm}

\usepackage{algorithm}
\usepackage{algorithmic}

\usepackage{color}

\newcommand{\mb}[1]{\mathbf{#1}}

\newcommand{\mc}[1]{\mathcal{#1}}
\renewcommand{\S}{\mathbf{S}}

\begin{document}

\title{Tensor Product Generation Networks for Deep NLP Modeling}
\author{Qiuyuan Huang, Paul Smolensky, Xiaodong He, Li Deng, Dapeng
Wu\\
\em{\{qihua,psmo,xiaohe\}@microsoft.com, l.deng@ieee.org,
dpwu@ufl.edu}\\ Microsoft Research AI\\
Redmond, WA
\thanks{This work was carried out while PS was on leave from Johns
Hopkins University. LD is currently at Citadel.  DW is with
University of Florida, Gainesville, FL 32611.} } \maketitle

\date{}

\begin{abstract}
We present a new approach to the design of deep networks for
natural language processing (NLP), based on the general technique
of Tensor Product Representations (TPRs) for encoding and
processing symbol structures in distributed neural networks. A
network architecture
--- the \emph{\textbf{Tensor Product Generation Network}} (\emph{\textbf{TPGN}}) --- is
proposed which is capable in principle of carrying out TPR
computation, but which uses unconstrained deep learning to design
its internal representations. Instantiated in a model for
image-caption generation, TPGN outperforms LSTM baselines when
evaluated on the COCO dataset. The TPR-capable structure enables
interpretation of internal representations and operations, which
prove to contain considerable grammatical content. Our
caption-generation model can be interpreted as generating
sequences of grammatical categories and retrieving words by their
categories from a plan encoded as a distributed representation.
\end{abstract}

\section{Introduction}
\label{sec:Introduction}

In this paper we introduce a new architecture for natural language
processing (NLP). On what type of principles can a computational
architecture be founded? It would seem a sound principle to
require that the hypothesis space for learning which an
architecture provides include network hypotheses that are
independently known to be suitable for performing the target task.
Our proposed architecture makes available to deep learning network
configurations that perform natural language generation by use of
\emph{Tensor Product Representations} (TPRs)
\cite{smolensky2006harmonic}. Whether learning will create TPRs is
unknown in advance, but what we can say with certainty is that the
hypothesis space being searched during learning includes TPRs as
one appropriate solution to the problem.

TPRs are a general method for generating vector-space embeddings of complex symbol structures.
Prior work has proved that TPRs enable powerful symbol processing to be carried out using neural network computation \cite{smolensky2012symbolic}.
This includes generating parse trees that conform to a grammar \cite{cho2017incremental}, although incorporating such capabilities into deep learning networks such as those developed here remains for future work.
The architecture presented here relies on simpler use of TPRs to generate sentences; grammars are not explicitly encoded here.

We test the proposed architecture by applying it to image-caption generation (on the MS-COCO dataset, \cite{COCO_weblink}).
The results improve upon a baseline deploying a state-of-the-art LSTM architecture \cite{vinyals2015show}, and the TPR foundations of the architecture provide greater interpretability.

Section~\ref{sec:ReviewTPR} of the paper reviews TPR.
Section~\ref{sec:GenArch} presents the proposed architecture, the
\emph{Tensor Product Generation Network} (TPGN).
Section~\ref{sec:System_Description} describes the particular
model we study for image captioning, and
Section~\ref{sec:ExperimentalResults} presents the experimental
results. Importantly, what the model has learned is interpreted in
Section~\ref{subsec:Interpretation}. Section~\ref{sec:RelatedWork}
discusses the relation of the new model to previous work and
Section~\ref{sec:Conclusion} concludes.

\section{Review of tensor product representation}
\label{sec:ReviewTPR}

The central idea of TPRs \cite{smolensky1990tensor} can be appreciated by contrasting the TPR for a word string with a bag-of-words (BoW) vector-space embedding.
In a BoW embedding, the vector that encodes \texttt{Jay saw Kay} is the same as the one that encodes \texttt{Kay saw Jay}: $\mb{J} + \mb{K} + \mb{s}$ where
$\mb{J}, \mb{K}, \mb{s}$ are respectively the vector embeddings of the words \texttt{Jay}, \texttt{Kay}, \texttt{saw}.

A TPR embedding that avoids this confusion starts by analyzing \texttt{Jay saw Kay} as the set \{\texttt{Jay}/{\sc subj}, \texttt{Kay}/{\sc obj}, \texttt{saw}/{\sc verb}\}.
(Other analyses are possible: see Section \ref{sec:GenArch}.)
Next we choose an embedding in a vector space $V_{F}$ for \texttt{Jay}, \texttt{Kay}, \texttt{saw} as in the BoW case: $\mb{J}, \mb{K}, \mb{s}$.
Then comes the step unique to TPRs: we choose an embedding in a vector space $V_{R}$ for the \emph{\textbf{roles}} {\sc subj}, {\sc obj}, {\sc verb}: $\mb{r}_{\textrm{\sc subj}}$, $\mb{r}_{\textrm{\sc obj}}$, $\mb{r}_{\textrm{\sc verb}}$.
Crucially, $\mb{r}_{\textrm{\sc subj}} \neq \mb{r}_{\textrm{\sc obj}}$.
Finally, the TPR for \texttt{Jay saw Kay} is the following vector in $V_{F} \otimes V_{R}$:
\begin{equation}
\label{eq:TPR1}
\mb{v_{\texttt{Jay saw Kay}}} = \mb{J} \otimes \mb{r}_{\textrm{\sc subj}} + \mb{K} \otimes \mb{r}_{\textrm{\sc obj}} + \mb{s} \otimes \mb{r}_{\textrm{\sc verb}}
\end{equation}
Each word is tagged with the role it fills in the sentence; \texttt{Jay} and \texttt{Kay} fill different roles.

This TPR avoids the BoW confusion: $\mb{v_{\texttt{Jay saw Kay}}} \neq \mb{v_{\texttt{Kay saw Jay}}}$ because
$\mb{J} \otimes \mb{r}_{\textrm{\sc subj}} + \mb{K} \otimes \mb{r}_{\textrm{\sc obj}} \neq
\mb{J}~\otimes~\mb{r}_{\textrm{\sc obj}} + \mb{K} \otimes \mb{r}_{\textrm{\sc subj}}$.
In the terminology of TPRs, in \texttt{Jay saw Kay}, \texttt{Jay} is the \emph{filler} of the role {\sc subj}, and $\mb{J} \otimes \mb{r}_{\textrm{\sc subj}}$ is the vector embedding of the \emph{filler/role binding} \texttt{Jay}/{\sc subj}.
In the vector space embedding, the binding operation is the tensor --- or generalized outer --- product $\otimes$;
i.e., $\mb{J} \otimes \mb{r}_{\textrm{\sc subj}}$ is a tensor with 2 indices defined by: $[\mb{J} \otimes \mb{r}_{\textrm{\sc subj}}]_{\varphi  \rho} \equiv [\mb{J}]_{\varphi} [\mb{r}_{\textrm{\sc subj}}]_{\rho}$.

The tensor product can be used recursively, which is essential for the TPR embedding of recursive structures such as trees and for the computation of recursive functions over TPRs.
However, in the present context, recursion will not be required, in which case the tensor product can be regarded as simply the matrix outer product (which cannot be used recursively); we can regard $\mb{J} \otimes \mb{r}_{\textrm{\sc subj}}$ as the matrix product $\mb{J} \mb{r}_{\textrm{\sc subj}}^{\top}$.
Then Equation \ref{eq:TPR1} becomes
\begin{equation}
\label{eq:TPR2}
\mb{v_{\texttt{Jay saw Kay}}} = \mb{J} \mb{r}_{\textrm{\sc subj}}^{\top} + \mb{K} \mb{r}_{\textrm{\sc obj}}^{\top} + \mb{s} \mb{r}_{\textrm{\sc verb}}^{\top}
\end{equation}
Note that the set of matrices (or the set of tensors with any fixed number of indices) is a vector space; thus \texttt{Jay saw Kay} $\mapsto \mb{v_{\texttt{Jay saw Kay}}}$ is a vector-space embedding of the symbol structures constituting sentences.
Whether we regard $\mb{v_{\texttt{Jay saw Kay}}}$ as a 2-index tensor or as a matrix, we can call it simply a `vector' since it is an element of a vector space:
in the context of TPRs, `vector' is used in a general sense and should not be taken to imply a single-indexed array.

Crucial to the computational power of TPRs and to the architecture we propose here is the notion of \emph{unbinding}.
Just as an \emph{outer} product --- the tensor product --- can be used to \emph{bind} the vector embedding a filler \texttt{Jay} to the vector embedding a role {\sc subj}, $\mb{J} \otimes \mb{r}_{\textrm{\sc subj}}$ or $\mb{J} \mb{r}_{\textrm{\sc subj}}^{\top}$, so an \emph{inner} product can be used to take the vector embedding a structure and \emph{unbind} a role contained within that structure, yielding the symbol that fills the role.

In the simplest case of orthonormal role vectors $\mb{r}_{i}$, to unbind role {\sc subj} in \texttt{Jay saw Kay} we can compute the matrix-vector product: $\mb{v_{\texttt{Jay saw Kay}}} \mb{r}_{\textrm{\sc subj}} = \mb{J}$ (because $\mb{r}_{i}^{\top} \mb{r}_{j} = \delta_{ij}$ when the role vectors are orthonormal).
A similar situation obtains when the role vectors are not orthonormal, provided they are not linearly dependent: for each role such as {\sc subj} there is an \emph{unbinding vector} $\mb{u}_{\textrm{\sc subj}}$ such that $\mb{r}_{i}^{\top} \mb{u}_{j} = \delta_{ij}$ so we get: $\mb{v_{\texttt{Jay saw Kay}}} \mb{u}_{\textrm{\sc subj}} = \mb{J}$.
A role vector such as $\mb{r}_{\textrm{\sc subj}}$ and its unbinding vector $\mb{u}_{\textrm{\sc subj}}$ are said to be \emph{duals} of each other.
(If $R$ is the matrix in which each column is a role vector $\mb{r}_{j}$, then $R$ is invertible when the role vectors are linearly independent; then the unbinding vectors $\mb{u}_{i}$ are the rows of $R^{-1}$.
When the $\mb{r}_{j}$ are orthonormal, $\mb{u}_{i} = \mb{r}_{i}$.
Replacing the matrix inverse with the pseudo-inverse allows approximate unbinding if the role vectors are linearly dependent.)

We can now see how TPRs can be used to generate a sentence one word at a time.
We start with the TPR for the sentence, e.g., $\mb{v_{\texttt{Jay saw Kay}}}$.
From this vector we unbind the role of the first word, which is {\sc subj}: the embedding of the first word is thus $\mb{v_{\texttt{Jay saw Kay}}} \mb{u}_{\textrm{\sc subj}} = \mb{J}$, the embedding of \texttt{Jay}.
Next we take the TPR for the sentence and unbind the role of the second word, which is {\sc verb}: the embedding of the second word is then
$\mb{v_{\texttt{Jay saw Kay}}} \mb{u}_{\textrm{\sc verb}} = \mb{s}$, the embedding of \texttt{saw}.
And so on.

\begin{figure*}[tbh]
    \centering
    \includegraphics[width=0.9\textwidth]{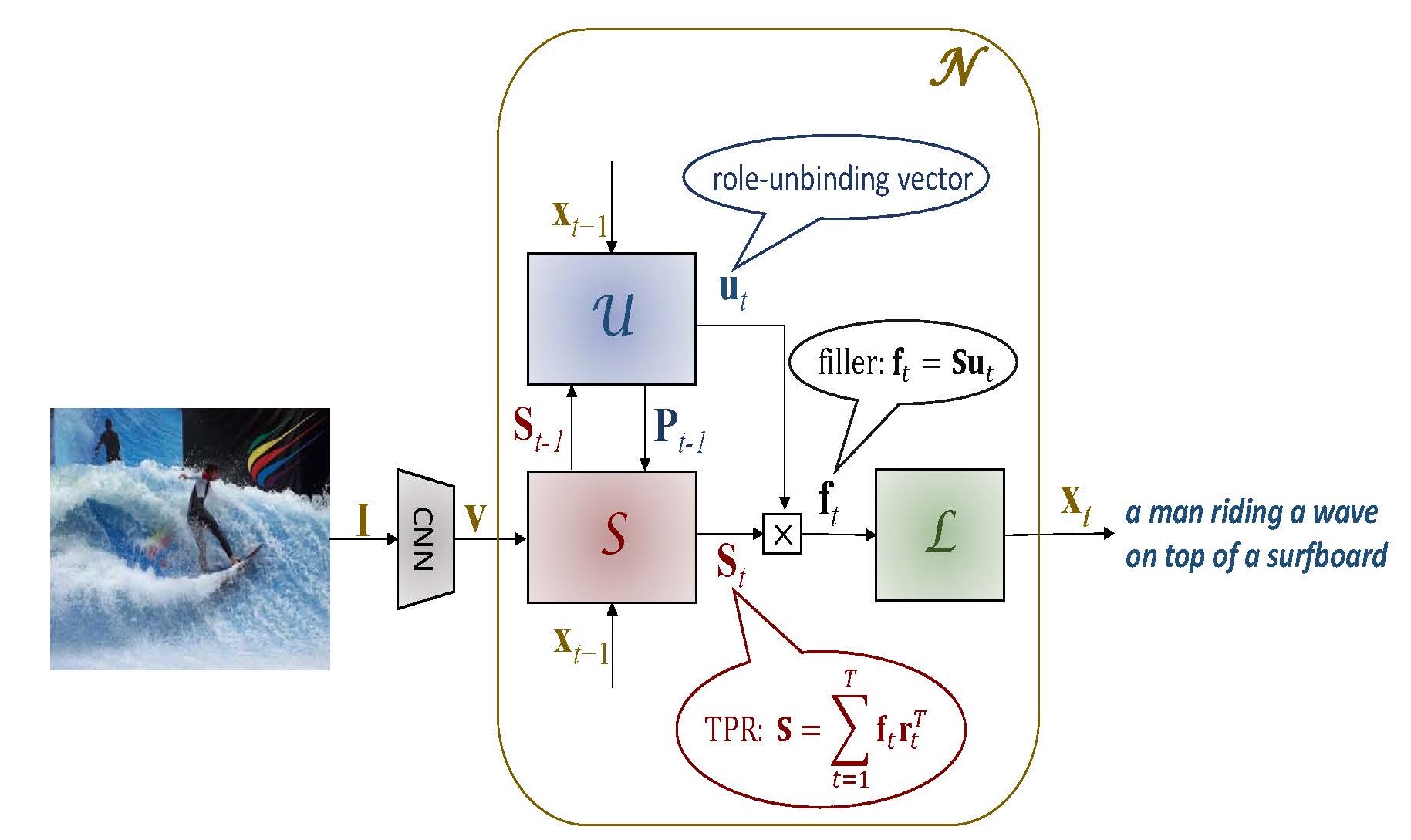}
    \caption{Architecture of TPGN, a TPR-capable generation network. ``$\Box\hspace{-8pt}\times$'' denotes the matrix-vector product.}
    \label{fig:Architecture1}
\end{figure*}

To accomplish this, we need two representations to generate the $t^{\textrm{th}}$ word: (i) the TPR of the sentence, $\mb{S}$ (or of the string of not-yet-produced words, $\mb{S}_{t}$) and (ii) the unbinding vector for the $t^{\textrm{th}}$ word, $\mb{u}_{t}$.
The architecture we propose will therefore be a recurrent network containing two subnetworks: (i) a subnet $\mc{S}$ hosting the representation $\mb{S}_{t}$, and a (ii) a subnet $\mc{U}$ hosting the unbinding vector $\mb{u}_{t}$.
This is shown in Fig. \ref{fig:Architecture1}.



\section{A TPR-capable generation architecture}
\label{sec:GenArch}

As Fig. \ref{fig:Architecture1} shows, the proposed Tensor Product Generation Network architecture (the dashed box labeled $\mc{N}$) is designed to support the technique for generation just described: the architecture is \emph{TPR-capable}.
There is a \emph{sentence-encoding subnetwork} $\mc{S}$ which \emph{could} host a TPR of the sentence to be generated, and an \emph{unbinding subnetwork} $\mc{U}$ which \emph{could} output a sequence of unbinding vectors $\mb{u}_{t}$; at time $t$, the embedding $\mb{f}_{t}$ of the word produced, $\texttt{x}_{t}$, could then be extracted from $\mb{S}_{t}$ via the matrix-vector product (shown in the figure by ``$\Box\hspace{-8pt}\times$''): $\mb{f}_{t} = \mb{S}_{t} \mb{u}_{t}$.
The lexical-decoding subnetwork $\mc{L}$ converts the embedding vector $\mb{f}_{t}$ to the 1-hot vector $\mb{x}_{t}$ corresponding to the word $\texttt{x}_{t}$.

Unlike some other work \cite{palangi2017deep}, TPGN is not constrained to literally learn TPRs.
The representations that will actually be housed in $\mc{S}$ and $\mc{U}$ are determined by end-to-end deep learning on a task:
the bubbles in Fig. \ref{fig:Architecture1} show what \emph{would} be the meanings of $\mb{S}_{t}, \mb{u}_{t}$ and $\mb{f_{t}}$ if an actual TPR scheme were instantiated in the architecture.
The learned representations $\S_{t}$ will not be proven to literally be TPRs, but by analyzing the unbinding vectors $\mb{u}_{t}$ the network learns, we will gain insight into the process by which the learned matrices $\S_{t}$ give rise to the generated sentence.

The task studied here is image captioning; Fig. \ref{fig:Architecture1} shows that the input to this TPGN model is an image, preprocessed by a CNN which produces the initial representation in $\mc{S}$, $\mb{S}_{0}$.
This vector $\mb{S}_{0}$ drives the entire caption-generation process: it contains all the image-specific information for producing the caption.
(We will call a caption a ``sentence''  even though it may in fact be just a noun phrase.)

The two subnets $\mc{S}$ and $\mc{U}$ are mutually-connected LSTMs \cite{hochreiter1997long}: see Fig. \ref{fig:Architecture2}.
The internal hidden state of $\mc{U}$, $\mb{p}_{t}$, is sent as input to $\mc{S}$; $\mc{U}$ also produces output, the unbinding vector $\mb{u}_{t}$.
The internal hidden state of $\mc{S}$, $\mb{S}_{t}$, is sent as input to $\mc{U}$, and also produced as output.
As stated above, these two outputs are multiplied together to produce the embedding vector $\mb{f}_{t} = \mb{S}_{t} \mb{u}_{t}$ of the output word $\texttt{x}_{t}$.
Furthermore, the 1-hot encoding $\mb{x}_{t}$ of $\texttt{x}_{t}$ is fed back at the next time step to serve as input to both $\mc{S}$ and $\mc{U}$.

What type of roles might the unbinding vectors be unbinding? A TPR for a caption could in principle be built upon \emph{positional roles}, \emph{syntactic/semantic roles}, or some combination of the two.
In the caption \texttt{a man standing in a room with a suitcase}, the initial \texttt{a} and \texttt{man} might respectively occupy the positional roles of
$\textsc{pos(ition)}_{1}$ and  $\textsc{pos}_{2}$; \texttt{standing} might occupy the syntactic role of \textsc{verb}; \texttt{in} the role of \textsc{Spatial-P(reposition)}; while \texttt{a room with a suitcase} might fill a 5-role schema $\textsc{Det(erminer)}_{1} \textsc{ N(oun)}_{1} \textsc{ P Det}_{2} \textsc{ N}_{2}$.
In fact we will provide evidence in Sec. \ref{subsubsec:clustering} that our network learns just this kind of hybrid role decomposition; further evidence for these particular roles is presented elsewhere.

What form of information does the sentence-encoding subnetwork $\mc{S}$ need to encode in $\S$?
Continuing with the example of the previous paragraph, $\S$ needs to be some approximation to the TPR summing several filler/role binding matrices.
In one of these bindings, a filler vector $\mb{f}_{\texttt{a}}$ --- which the lexical subnetwork $\mc{L}$ will map to the article \texttt{a} --- is bound (via the outer product) to a role vector $\mb{r}_{\textsc{Pos}_{1}}$ which is the dual of the first unbinding vector produced by the unbinding subnetwork $\mc{U}$: $\mb{u}_{\textsc{Pos}_{1}}$.
In the first iteration of generation the model computes $\S_{1} \mb{u}_{\textsc{Pos}_{1}} = \mb{f}_{\texttt{a}}$, which $\mc{L}$ then maps to \texttt{a}.
Analogously, another binding approximately contained in $\S_{2}$ is $\mb{f}_{\texttt{man}} \mb{r}_{\textsc{Pos}_{2}}^{\top}$.
There are corresponding approximate bindings for the remaining words of the caption; these employ syntactic/semantic roles.
One example is $\mb{f}_{\texttt{standing}} \mb{r}_{V}^{\top}$.
At iteration 3, $\mc{U}$ decides the next word should be a verb, so it generates the unbinding vector $\mb{u}_{V}$ which when multiplied by the
current output of $\mc{S}$, the matrix $\S_{3}$, yields a filler vector $\mb{f}_{\texttt{standing}}$ which $\mc{L}$ maps to the output \texttt{standing}.
$\mc{S}$ decided the caption should deploy \texttt{standing} as a verb and included in $\S$ an approximation to the binding $\mb{f}_{\texttt{standing}} \mb{r}_{V}^{\top}$.
It similarly decided the caption should deploy \texttt{in} as a spatial preposition, approximately including in $\S$ the binding $\mb{f}_{\texttt{in}}
\mb{r}_{\textsc{Spatial-P}}^{\top}$; and so on for the other words in their respective roles in the caption.

\section{System Description}
\label{sec:System_Description}

\begin{figure*}[htb]
    \centering
    \includegraphics[width=0.7\textwidth]{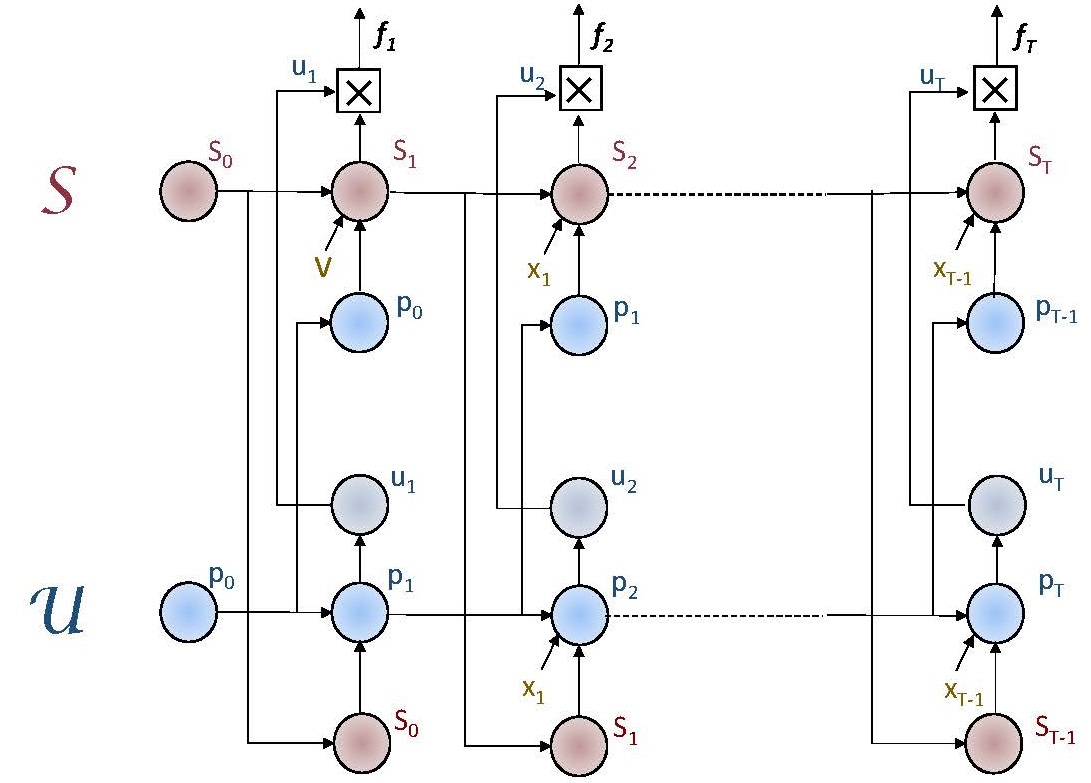}
    \caption{The sentence-encoding subnet $\mc{S}$ and the unbinding subnet $\mc{U}$ are inter-connected LSTMs; $\mb{v}$ encodes the visual input while the $\mb{x}_{t}$ encode the words of the output caption.
    }
    \label{fig:Architecture2}
\end{figure*}

As stated above, the unbinding subnetwork $\mc{U}$ and the sentence-encoding subnetwork $\mc{S}$ of Fig. \ref{fig:Architecture1} are each
implemented as (1-layer, 1-directional) LSTMs (see Fig. \ref{fig:Architecture2}); the lexical subnetwork $\mc{L}$ is implemented as a linear transformation followed by a softmax operation.

In the equations below, the LSTM variables internal to the $\mc{S}$ subnet are indexed by 1 (e.g., the forget-, input-, and output-gates are respectively $\mb{\hat{f}}_{1}, \mb{\hat{i}}_{1}, \mb{\hat{o}}_{1}$) while those of the unbinding subnet $\mc{U}$ are indexed by 2.

Thus the state updating equations for $\mc{S}$ are, for
$t=1,\cdots,T$ = caption length: {\small
\begin{eqnarray}
{\mathbf {\hat{f}}}_{1,t}=\sigma_g({\mathbf W}_{1,f} {\mathbf
p}_{t-1}-{\mathbf D}_{1,f}{\mathbf W}_e{\mathbf x}_{t-1}+{\mathbf
U}_{1,f}{\mathbf {\hat{S}}}_{t-1})\label{eq:TPR-LSTM1_1}\\
{\mathbf {\hat{i}}}_{1,t}=\sigma_g({\mathbf W}_{1,i} {\mathbf
p}_{t-1}-{\mathbf D}_{1,i}{\mathbf W}_e {\mathbf x}_{t-1}+{\mathbf
U}_{1,i}{\mathbf {\hat{S}}}_{t-1})\\
{\mathbf {\hat{o}}}_{1,t}=\sigma_g({\mathbf W}_{1,o} {\mathbf
p}_{t-1}-{\mathbf D}_{1,o}{\mathbf W}_e {\mathbf x}_{t-1}+{\mathbf
U}_{1,o}{\mathbf {\hat{S}}}_{t-1})\\
{\mathbf g}_{1,t}=\sigma_h({\mathbf W}_{1,c} {\mathbf
p}_{t-1}-{\mathbf D}_{1,c}{\mathbf W}_e {\mathbf x}_{t-1}+{\mathbf
U}_{1,c}{\mathbf {\hat{S}}}_{t-1})\\
{\mathbf c}_{1,t}= {\mathbf {\hat{f}}}_{1,t} \odot {\mathbf
c}_{1,t-1}+
{\mathbf {\hat{i}}}_{1,t} \odot {\mathbf g}_{1,t} \quad\quad\quad\\
 {\mathbf {\hat{S}}}_t= {\mathbf {\hat{o}}}_{1,t} \odot \sigma_h({\mathbf
 c}_{1,t})\quad\quad\quad\quad\quad\quad\quad
\label{eq:TPR-LSTM1_5}
\end{eqnarray}    }
Here ${\mathbf {\hat{f}}}_{1,t}$, ${\mathbf {\hat{i}}}_{1,t}$,
${\mathbf {\hat{o}}}_{1,t}$, ${\mathbf g}_{1,t}$, ${\mathbf
c}_{1,t}$, ${\mathbf {\hat{S}}}_t\in \mathbb{R}^{d\times d}$,
${\mathbf p}_t\in \mathbb{R}^{d}$; $\sigma_g(\cdot)$ is the
(element-wise) logistic sigmoid function; $\sigma_h(\cdot)$ is the
hyperbolic tangent function; the operator $\odot$ denotes the
Hadamard (element-wise) product; ${\mathbf W}_{1,f},{\mathbf
W}_{1,i},{\mathbf W}_{1,o},{\mathbf W}_{1,c} \in
\mathbb{R}^{d\times d \times d}$, ${\mathbf D}_{1,f}$, $ {\mathbf
D}_{1,i}$, $ {\mathbf D}_{1,o}$, $ {\mathbf D}_{1,c}$ $\in$
$\mathbb{R}^{d\times d \times d}$, ${\mathbf U}_{1,f}$, $ {\mathbf
U}_{1,i}$, $ {\mathbf U}_{1,o}$, $ {\mathbf U}_{1,c}$ $\in$
$\mathbb{R}^{d\times d \times d\times d}$. For clarity, biases ---
included throughout the model --- are omitted from all equations
in this paper. The initial state ${\mathbf {\hat{S}}}_0$ is
initialized by:
\begin{equation}
{\mathbf {\hat{S}}}_0={\mathbf C}_s ({\mathbf v}-\bar{{\mathbf
v}})
 \label{eq:TPR-RNN1_4}
\end{equation}
where ${\mathbf v} \in \mathbb{R}^{2048}$ is the vector of visual
features extracted from the current image by ResNet
\cite{SCN_CVPR2017} and $\bar{{\mathbf v}}$ is the mean of all
such vectors;
 ${\mathbf C}_s \in
\mathbb{R}^{d\times d \times 2048}$. On the output side, ${\mathbf
x}_t \in \mathbb{R}^{V}$ is a 1-hot vector with dimension equal to
the size of the caption vocabulary, $V$, and ${\mathbf W}_e \in
\mathbb{R}^{d\times V}$ is a word embedding matrix, the $i$-th
column of which is the embedding vector of the $i$-th word in the
vocabulary; it is obtained by the Stanford GLoVe algorithm with
zero mean \cite{Stanford_Glove_weblink}. ${\mathbf x}_0$ is
initialized as the one-hot vector corresponding to a
 ``start-of-sentence'' symbol.

For $\mc{U}$ in Fig.~\ref{fig:Architecture1}, the state updating
equations are:{\small
\begin{eqnarray}
{\mathbf {\hat{f}}}_{2,t}=\sigma_g({\mathbf
{\hat{S}}}_{t-1}{\mathbf w}_{2,f}-{\mathbf D}_{2,f}{\mathbf W}_e
{\mathbf
x}_{t-1}+{\mathbf U}_{2,f}{\mathbf p}_{t-1})\label{eq:TPR-LSTM2_1}\\
{\mathbf {\hat{i}}}_{2,t}=\sigma_g({\mathbf
{\hat{S}}}_{t-1}{\mathbf w}_{2,i}-{\mathbf D}_{2,i}{\mathbf W}_e
{\mathbf x}_{t-1}+{\mathbf
U}_{2,i}{\mathbf p}_{t-1})\\
{\mathbf {\hat{o}}}_{2,t}=\sigma_g({\mathbf
{\hat{S}}}_{t-1}{\mathbf w}_{2,o}-{\mathbf D}_{2,o}{\mathbf W}_e
{\mathbf x}_{t-1}+{\mathbf
U}_{2,o}{\mathbf p}_{t-1})\\
{\mathbf g}_{2,t}=\sigma_h({\mathbf {\hat{S}}}_{t-1}{\mathbf
w}_{2,c}-{\mathbf D}_{2,c}{\mathbf W}_e {\mathbf x}_{t-1}+{\mathbf
U}_{2,c}{\mathbf p}_{t-1})\\
 {\mathbf c}_{2,t}= {\mathbf {\hat{f}}}_{2,t} \odot
{\mathbf c}_{2,t-1}+ {\mathbf {\hat{i}}}_{2,t} \odot {\mathbf g}_{2,t}\quad\quad\quad\\
{\mathbf p}_t={\mathbf {\hat{o}}}_{2,t} \odot \sigma_h({\mathbf
c}_{2,t})\quad\quad\quad\quad\quad\quad\quad
 \label{eq:TPR-LSTM2_5}
\end{eqnarray}   }
Here ${\mathbf w}_{2,f}, {\mathbf w}_{2,i}, {\mathbf w}_{2,o}, {\mathbf w}_{2,c} \in \mathbb{R}^{d}$,
${\mathbf D}_{2,f}$, ${\mathbf
D}_{2,i}$, ${\mathbf D}_{2,o}$, ${\mathbf D}_{2,c}$ $\in$ $\mathbb{R}^{d
\times d}$, and ${\mathbf U}_{2,f}$, ${\mathbf U}_{2,i}$, ${\mathbf
U}_{2,o}$, ${\mathbf U}_{2,c}$ $\in$ $\mathbb{R}^{d\times d}$. The
initial state ${\mathbf p}_0$ is the zero vector.

The dimensionality of the crucial vectors shown in Fig.
\ref{fig:Architecture1}, $\mb{u}_{t}$ and $\mb{f}_{t}$, is
increased from $d \times 1$ to $d^{2} \times 1$ as follows. A
block-diagonal $d^{2} \times d^{2}$ matrix $\mb{S}_{t}$ is created
by placing $d$ copies of the $d \times d$ matrix
$\mb{\hat{S}}_{t}$ as blocks along the principal diagonal. This
matrix is the output of the sentence-encoding subnetwork $\mc{S}$.
Now the  `filler vector'
${\mathbf f}_t\in \mathbb{R}^{d^{2}}$ --- `unbound' from the
sentence representation $\S_{t}$ with the `unbinding vector'
$\mb{u}_{t}$ --- is obtained by Eq. \eqref{eq:TPR-RNN1f}.
\begin{equation}
{\mathbf f}_t= {\mathbf S}_t  {\mathbf u}_t
 \label{eq:TPR-RNN1f}
\end{equation}
Here ${\mathbf u}_t\in \mathbb{R}^{d^{2}}$, the output of the
unbinding subnetwork $\mc{U}$, is computed as in Eq.
\eqref{eq:TPR-RNN2u}, where ${\mathbf W}_u \in
\mathbb{R}^{d^{2}\times d}$ is $\mc{U}$'s output weight matrix.
\begin{equation}
{\mathbf u}_t=\sigma_h({\mathbf W}_u{\mathbf p}_t)
 \label{eq:TPR-RNN2u}
\end{equation}

Finally, the lexical subnetwork $\mc{L}$ produces a decoded word
${\mathbf x}_t\in \mathbb{R}^{V}$ by
\begin{equation}
{\mathbf x}_t=\sigma_s({\mathbf W}_x{\mathbf f}_t)
 \label{eq:TPR-FFNN}
\end{equation}
where $\sigma_s(\cdot)$ is the softmax function and ${\mathbf W}_x
\in \mathbb{R}^{V\times d^2}$ is the overall output weight matrix.
Since ${\mathbf W}_x$ plays the role of a word de-embedding
matrix, we can set
\begin{equation}
{\mathbf W}_x=({\mathbf W}_e)^\top
 \label{eq:embeddingVector2}
\end{equation}
where ${\mathbf W}_e$ is the word-embedding matrix. Since
${\mathbf W}_e$ is pre-defined, we directly set ${\mathbf W}_x$ by
Eq.~\eqref{eq:embeddingVector2} without training $\mc{L}$ through
Eq.~\eqref{eq:TPR-FFNN}. Note that $\mc{S}$ and $\mc{U}$ are
learned jointly through the end-to-end training as shown in
Algorithm~\ref{alg:trainingSLSTM}.

\begin{algorithm}[htb]
\caption{End-to-end training of $\mc{S}$ and $\mc{U}$}
\label{alg:trainingSLSTM} \textbf{Input:} Image feature vector
${\mathbf v}^{(i)}$ and corresponding
caption ${\mathbf X}^{(i)}=[{\mathbf x}_1^{(i)}$, $\cdots $, $ {\mathbf x}_T^{(i)}]$ ($i = 1$ , $\cdots$, $N$), where $N$ is the total number of samples.\\
\textbf{Output:}  ${\mathbf W}_{1,f},{\mathbf W}_{1,i},{\mathbf
W}_{1,o},{\mathbf W}_{1,c}, {\mathbf C}_{s}, {\mathbf D}_{1,f},
{\mathbf D}_{1,i},{\mathbf D}_{1,o}$,\\ ${\mathbf D}_{1,c},
{\mathbf U}_{1,f}, {\mathbf U}_{1,i}, {\mathbf U}_{1,o}, {\mathbf
U}_{1,c}, {\mathbf w}_{2,f}, {\mathbf w}_{2,i}, {\mathbf w}_{2,o},
 {\mathbf w}_{2,c}, {\mathbf D}_{2,f}$,\\ ${\mathbf
D}_{2,i},{\mathbf D}_{2,o},{\mathbf D}_{2,c}, {\mathbf U}_{2,f},
{\mathbf U}_{2,i}, {\mathbf U}_{2,o}, {\mathbf U}_{2,c}, {\mathbf
W}_u, {\mathbf W}_x$.
\begin{algorithmic}[1]
\STATE Initialize ${\mathbf S}_0$ by \eqref{eq:TPR-RNN1_4}; \STATE
Initialize ${\mathbf x}_0$ as the one-hot vector corresponding to
the start-of-sentence symbol; \STATE Initialize ${\mathbf p}_0$ as
the zero vector; \STATE Randomly initialize weights {\small
${\mathbf W}_{1,f},{\mathbf W}_{1,i},{\mathbf W}_{1,o}$,\\
${\mathbf W}_{1,c}, {\mathbf C}_{s},{\mathbf D}_{1,f}, {\mathbf
D}_{1,i}, {\mathbf D}_{1,o}, {\mathbf D}_{1,c}, {\mathbf U}_{1,f},
{\mathbf U}_{1,i}, {\mathbf U}_{1,o}$,\\ ${\mathbf U}_{1,c},
{\mathbf w}_{2,f}, {\mathbf w}_{2,i}, {\mathbf w}_{2,o}, {\mathbf
w}_{2,c}, {\mathbf D}_{2,f}, {\mathbf D}_{2,i}, {\mathbf
D}_{2,o}$,\\ ${\mathbf D}_{2,c}, {\mathbf U}_{2,f}, {\mathbf
U}_{2,i}, {\mathbf U}_{2,o}, {\mathbf U}_{2,c}, {\mathbf W}_u,
{\mathbf W}_x$;} \FOR {$n$ from $1$ to $N$}   \FOR {$t$ from $1$
to $T$} \STATE Calculate \eqref{eq:TPR-LSTM1_1} --
\eqref{eq:TPR-LSTM1_5} to obtain ${\mathbf S}_t$; \STATE Calculate
\eqref{eq:TPR-LSTM2_1} -- \eqref{eq:TPR-LSTM2_5} to obtain
${\mathbf p}_t$; \STATE Calculate \eqref{eq:TPR-RNN2u} to obtain
${\mathbf u}_t$; \STATE Calculate \eqref{eq:TPR-RNN1f} to obtain
${\mathbf f}_t$; \STATE Calculate \eqref{eq:TPR-FFNN} to obtain
${\mathbf x}_t$; \STATE Update weights {\small ${\mathbf
W}_{1,f},{\mathbf W}_{1,i},{\mathbf W}_{1,o}$,\\ ${\mathbf
W}_{1,c}, {\mathbf C}_{s}, {\mathbf D}_{1,f},  {\mathbf D}_{1,i},
{\mathbf D}_{1,o}, {\mathbf D}_{1,c}, {\mathbf U}_{1,f}, {\mathbf
U}_{1,i}$,\\   ${\mathbf U}_{1,o}, {\mathbf U}_{1,c}, {\mathbf
w}_{2,f}, {\mathbf w}_{2,i},{\mathbf w}_{2,o}, {\mathbf w}_{2,c},
{\mathbf D}_{2,f},{\mathbf D}_{2,i}$,\\ ${\mathbf
D}_{2,o},{\mathbf D}_{2,c}, {\mathbf U}_{2,f}, {\mathbf U}_{2,i},
{\mathbf U}_{2,o}, {\mathbf U}_{2,c}, {\mathbf W}_u, {\mathbf
W}_x$ }\\ by the back-propagation algorithm; \ENDFOR \ENDFOR
\end {algorithmic}
\end{algorithm}

\section{Experimental results}
\label{sec:ExperimentalResults}

\subsection{Dataset}\label{subsec:Dataset}
To evaluate the performance of our proposed model, we use the
COCO dataset \cite{COCO_weblink}. The COCO dataset contains
123,287 images, each of which is annotated with at least 5
captions. We use the same pre-defined splits as in
\cite{karpathy2015deep,SCN_CVPR2017}: 113,287 images for training,
5,000 images for validation, and 5,000 images for testing. We use
the same vocabulary as that employed in \cite{SCN_CVPR2017}, which
consists of 8,791 words.

\subsection{Evaluation}\label{subsec:Evaluation}
For the CNN of Fig.~\ref{fig:Architecture1}, we used ResNet-152
\cite{he2016deep}, pretrained on the ImageNet dataset. The feature
vector ${\mathbf v}$ has 2048 dimensions. Word embedding vectors
in ${\mathbf W}_e$ are downloaded from the web
\cite{Stanford_Glove_weblink}. The model is implemented in
TensorFlow \cite{tensorflow2015-whitepaper} with the default
settings for random initialization and optimization by
backpropagation.

In our experiments, we choose $d=25$ (where $d$ is the dimension
of vector ${\mathbf p}_t$).  The dimension of ${\mathbf S}_t$ is
 $625 \times 625$ (while $\mb{\hat{S}}_{t}$ is $25 \times 25$);
 the vocabulary size $V=8,791$; the dimension of ${\mathbf  u}_t$ and ${\mathbf f}_t$ is
 $d^2=625$.

{\small
\begin{table*}[htb]
  \caption{Performance of the proposed TPGN model on the COCO dataset.}
  \label{table:BLEU}
  \centering
  \begin{tabular}{lllllllll}
    \hline
Methods     & METEOR &BLEU-1 & BLEU-2 & BLEU-3 & BLEU-4 &   CIDEr\\
    \hline
NIC \cite{vinyals2015show} & 0.237 &  0.666& 0.461& 0.329& 0.246&  0.855\\
CNN-LSTM &0.238 & 0.698 &0.525 &0.390 &0.292  & 0.889 \\
 TPGN  & \textbf{0.243}& \textbf{0.709} & \textbf{0.539} &  \textbf{0.406} & \textbf{0.305}  &  \textbf{0.909}\\
    \hline
  \end{tabular}
\end{table*}
}

The main evaluation results on the MS COCO dataset are reported in
Table~\ref{table:BLEU}. The widely-used BLEU
\cite{papineni2002bleu}, METEOR  \cite{banerjee2005meteor}, and
CIDEr  \cite{vedantam2015cider} metrics are reported in our
quantitative evaluation of the performance of the proposed
model. In evaluation, our baseline is the widely used CNN-LSTM
captioning method originally proposed in \cite{vinyals2015show}.
For comparison, we include results in that paper in the first line
of Table~\ref{table:BLEU}. We also re-implemented the model using
the latest ResNet features and report the results in the second
line of Table~\ref{table:BLEU}. Our re-implementation of the
CNN-LSTM method matches the performance reported in
\cite{SCN_CVPR2017}, showing that the baseline is a
state-of-the-art implementation. As shown in
Table~\ref{table:BLEU}, compared to the CNN-LSTM baseline, the
proposed TPGN significantly outperforms the benchmark schemes in
all metrics across the board. The improvement in BLEU-$n$ is
greater for greater $n$; TPGN particularly improves generation of
longer subsequences. The results attest to the
effectiveness of the TPGN architecture.

\subsection{Interpretation of learned unbinding vectors}
\label{subsec:Interpretation}

To get a sense of how the sentence encodings $\S_{t}$  learned by
TPGN approximate TPRs, we now investigate the meaning of the
role-unbinding vector $\mb{u}_{t}$ the model uses to unbind from
$\S_{t}$ --- via Eq.~\eqref{eq:TPR-RNN1f} --- the filler vector
$\mb{f}_{t}$ that  produces --- via Eq.~\eqref{eq:TPR-FFNN} ---
the one-hot vector  $\mb{x}_{t}$ of the $t^\textrm{th}$ generated caption
word. The meaning of an unbinding vector is the meaning of the
role it unbinds.  Interpreting the unbinding vectors reveals the
meaning of the roles in a TPR that $\S$ approximates.

\begin{figure}[tbh]
    \centering
    \includegraphics[width=0.6\textwidth]{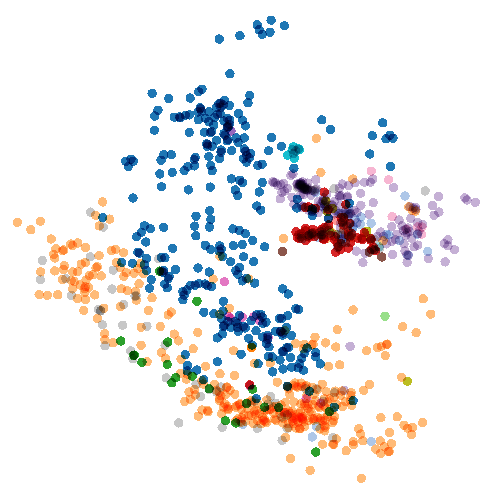}
    \caption{Unbinding vectors of 1000 words; different POS tags of words are represented by different colors.}
    \label{fig:total}
\end{figure}

\subsubsection{Visualization of ${\mathbf u}_t$}
\label{subsubsec:Visualization}

We run the TPGN model with 5,000 test images as input, and obtain
the unbinding vector ${\mathbf u}_t$ used to generate each word
${\mathbf x}_t$ in the caption of a test image. We plot 1,000
unbinding vectors ${\mathbf u}_t$, which correspond to the first
1,000 words in the resulting captions of these 5,000 test images.
There are 17 parts of speech (POS) in these 1,000 words. The POS
tags are obtained by the Stanford Parser
\cite{Stanford_Parser_weblink}.

\begin{figure}[tbh]
    \centering
    \includegraphics[width=0.6\textwidth]{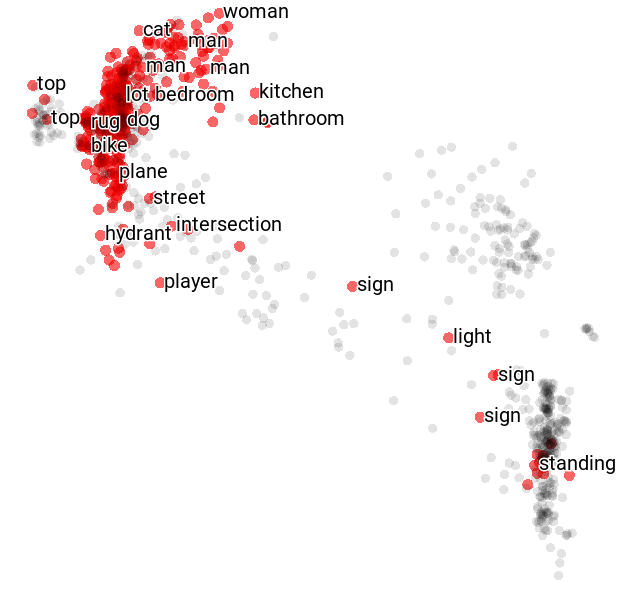}
    \caption{Unbinding vectors of 360 nouns in red and 640 words of other types of POS in grey.}
    \label{fig:nouns}
\end{figure}

\begin{figure}[tbh]
    \centering
    \includegraphics[width=0.6\textwidth]{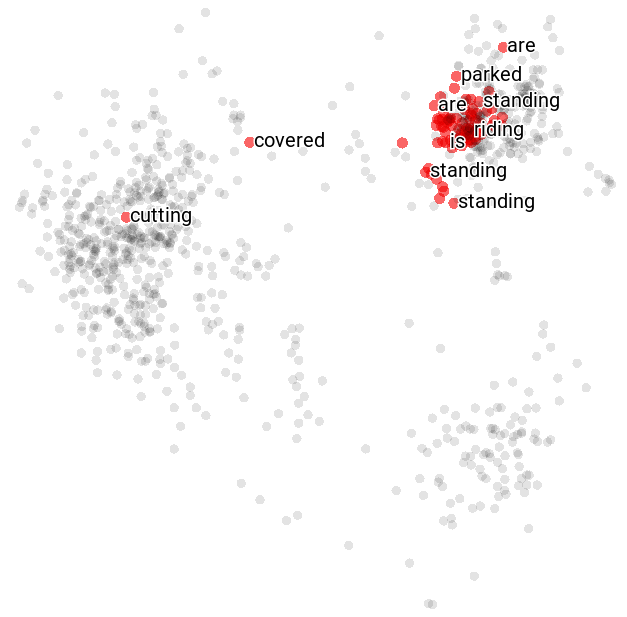}
    \caption{Unbinding vectors of 81 verbs in red and 919 words of other types of POS in grey.}
    \label{fig:verbs}
\end{figure}

\begin{figure}[tbh]
    \centering
    \includegraphics[width=0.6\textwidth]{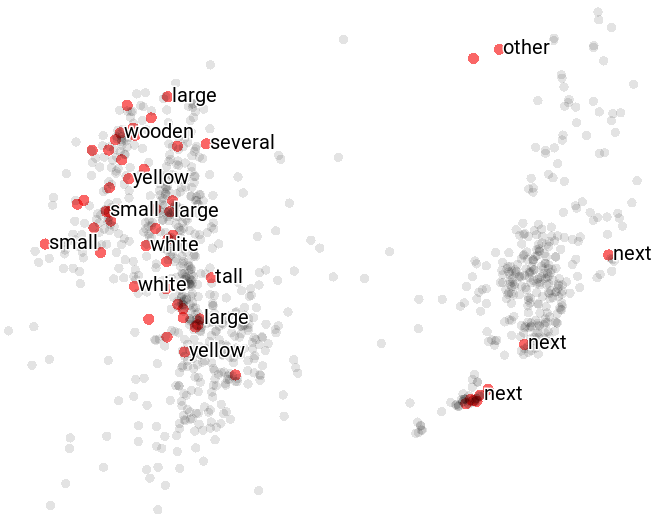}
    \caption{Unbinding vectors of 55 adjectives in red and 945 words of other types of POS in grey.}
    \label{fig:adjectives}
\end{figure}

\begin{figure}[tbh]
    \centering
    \includegraphics[width=0.6\textwidth]{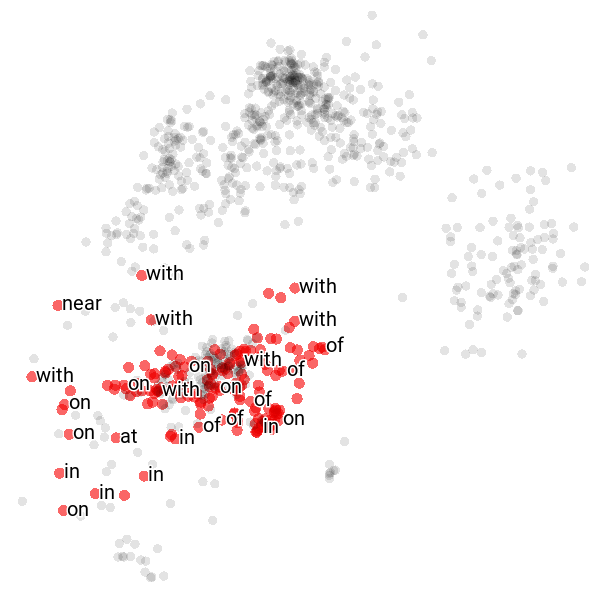}
    \caption{Unbinding vectors of 169 prepositions in red and 831 words of other types of POS in grey.}
    \label{fig:IN}
\end{figure}

We use the Embedding Projector in TensorBoard
\cite{Embeddings_weblink} to plot 1,000 unbinding vectors
${\mathbf u}_t$ with a custom linear projection in TensorBoard to
reduce 625 dimensions of ${\mathbf u}_t$ to 2 dimensions shown in
Fig.~\ref{fig:total} through Fig.~\ref{fig:IN}.

Fig.~\ref{fig:total} shows the unbinding vectors of 1000 words;
different POS tags of words are represented by different colors.
In fact, we can partition the 625-dim space of ${\mathbf u}_t$
into 17 regions, each of which contains 76.3\% words of the same
type of POS on average; i.e., each region is dominated by words of
one POS type. This clearly indicates that \emph{each unbinding
vector contains important grammatical information about the word
it generates}. As examples, Fig.~\ref{fig:nouns} to
Fig.~\ref{fig:IN} show the distribution of the unbinding vectors
of nouns, verbs, adjectives, and prepositions, respectively.

{\small
\begin{table}[htb]
  \caption{Conformity to N/V generalization ($N_u=2)$.}
  \label{table:Statistics2}
  \centering
{\small  \begin{tabular}{lrrr}  
    \hline
Category & $N_w$ & $N_r$ & $P_c$\\\hline\hline
Nouns &   16683 & 16115 & 0.969   \\
Pronouns & 462&442&   0.957  \\
Indefinite articles & 7248&  7107&  0.981 \\
Definite articles & 797&   762&  0.956 \\
Adjectives & 2543&  2237&  0.880 \\\hline
Verbs & 3558& 3409&  0.958 \\
Prepositions \& conjunctions & 8184 &7859 &  0.960 \\
 Adverbs & 13 &    8 & 0.615
 \\\hline
  \end{tabular}
  }
\end{table}
}


{\small
\begin{table}[htb]
  \caption{Interpretation of unbinding clusters ($N_u=10)$}
  \label{table:RoleClusters}
  \centering
{\small  \begin{tabular}{ll}
    \hline
ID  & Interpretation (proportion)\\\hline\hline
2 & Position 1 (1.00) \\
3 & Position 2 (1.00) \\
\hline
1 & Noun (0.54), Determiner (0.43)  \\
5 & Determiner (0.50), Noun (0.19), Preposition (0.15) \\
7 & Noun (0.88), Adjective (0.09) \\
9 & Determiner (0.90), Noun (0.10) \\
\hline
0 & Preposition (0.64), . (0.16), V (0.14) \\
4 & Preposition: spatial (0.72) non-spatial (0.19) \\
6 & Preposition (0.59), . (0.14) \\
8 & Verb (0.37), Preposition (0.36), . (0.20) \\
\hline
  \end{tabular}
  }
\end{table}
}

\subsubsection{Clustering of ${\mathbf u}_t$}
\label{subsubsec:clustering}

Since the previous section indicates that there is a clustering structure for ${\mathbf u}_t$, in this section we partition ${\mathbf u}_t$ into
$N_u$ clusters and examine the grammar roles played by ${\mathbf u}_t$ .

First, we run the trained TPGN model on the 113,287 training
images, obtaining the role-unbinding vector ${\mathbf u}_t$ used
to generate each word ${\mathbf x}_t$ in the caption sentence.
There are approximately 1.2 million ${\mathbf u}_t$ vectors over
all the training images. We apply the K-means clustering algorithm
to these vectors to obtain $N_u$ clusters and the centroid ${\bm \mu}_i$ of each cluster $i$
($i=0,\cdots,N_u-1$).

Then, we run the TPGN model with 5,000 test images as input, and
obtain the role vector ${\mathbf u}_t$ of each word ${\mathbf
x}_t$ in the caption sentence of a test image. Using the nearest
neighbor rule, we obtain the index $i$ of the cluster that each
${\mathbf u}_t$ is assigned to.

The partitioning of the unbinding vectors ${\mathbf u}_t$ into $N_{u} = 2$
clusters exposes the most fundamental distinction made by the
roles. We find that the vectors assigned to Cluster 1 generate
words which are nouns, pronouns, indefinite and definite
articles, and adjectives, while the vectors assigned to Cluster 0
generate verbs, prepositions, conjunctions, and adverbs. Thus
Cluster 1 contains the noun-related words, Cluster 0 the verb-like
words (verbs, prepositions and conjunctions are all potentially
followed by noun-phrase complements, for example). Cross-cutting
this distinction is another dimension, however: the initial word
in a caption (always a determiner) is sometimes generated with a
Cluster 1 unbinding vector, sometimes with a Cluster 0 vector.
Outside the caption-initial position, exceptions to the
nominal/verbal $\sim$ Cluster 1/0 generalization are rare, as
attested by the high rates of conformity to the generalization
shown in Table \ref{table:Statistics2} .

Table~\ref{table:Statistics2} shows the likelihood of correctness
of this `N/V' generalization for the words in 5,000 sentences
captioned for the 5,000 test images; $N_w$ is the number of words
in the category, $N_r$ is the number of words conforming to the
generalization, and $P_c=N_r/N_w$ is the proportion conforming. We
use the Natural Language Toolkit \cite{NLTK_weblink} to identify
the part of speech of each word in the captions.

A similar analysis with $N_{u} = 10$ clusters reveals the results shown in
Table~\ref{table:RoleClusters}; these results concern the first
100 captions, which were inspected manually to identify
interpretable patterns. (More comprehensive results will be
discussed elsewhere.)

The clusters can be interpreted as falling into 3 groups (see
Table~\ref{table:RoleClusters}). Clusters 2 and 3 are clearly
positional roles: every initial word is generated by a
role-unbinding vector from Cluster 2, and such vectors are not
used elsewhere in the string. The same holds for Cluster 3 and the
second caption word.

For caption words after the second word, position is replaced by
syntactic/semantic properties for interpretation purposes. The
 vector clusters aside from 2 and 3 generate words with a
dominant grammatical category: for example, unbinding vectors
assigned to the cluster 4 generate words that are 91\%
likely to be prepositions, and 72\% likely to be spatial
prepositions. Cluster 7 generates 88\% nouns and 9\% adjectives,
with the remaining 3\% scattered across other categories. As
Table~\ref{table:RoleClusters} shows, clusters 1, 5, 7, 9 are
primarily nominal, and 0, 4, 6, and 8 primarily verbal. (Only
cluster 5 spans the N/V divide.)


\section{Related work}
\label{sec:RelatedWork}

This work follows a great deal of recent caption-generation
literature in exploiting end-to-end deep learning with a CNN
image-analysis front end producing a distributed representation
that is then used to drive a natural-language generation process,
typically using RNNs
\cite{mao2015deep,vinyals2015show,devlin2015language,chen2015mind,donahue2015long,karpathy2015deep,kiros2014multimodal,kiros2014unifying}.
Our grammatical interpretation of the structural roles of words in
sentences makes contact with other work that incorporates deep
learning into grammatically-structured networks
\cite{tai2015improved,kumar2016ask,kong2017dragnn,andreas2015deep,yogatama2016learning,maillard2017jointly,socher2010learning,pollack1990recursive}.
Here, the network is not itself structured to match the
grammatical structure of sentences being processed; the structure
is fixed, but is designed to support the learning of distributed
representations that incorporate structure internal to the
representations themselves
--- filler/role structure.

TPRs are also used in NLP in \cite{palangi2017deep} but there the
representation of each individual input word is constrained to be
a literal TPR filler/role binding. (The idea of using the outer
product to construct internal representations was also explored in
\cite{fukui2016multimodal}.) Here, by contrast, the learned
representations are not themselves constrained, but the global
structure of the network is designed to display the somewhat
abstract property of being TPR-capable: the architecture uses the
TPR unbinding operation of the matrix-vector product to extract
individual words for sequential output.

\section{Conclusion}
\label{sec:Conclusion}
Tensor Product Representation (TPR) \cite{smolensky1990tensor} is a general technique for constructing vector embeddings of complex symbol structures in such a way that powerful symbolic functions can be computed using hand-designed neural network computation.
Integrating TPR with deep learning is a largely open problem for which the work presented here proposes a general approach: design deep architectures that are TPR-capable --- TPR computation is within the scope of the capabilities of the architecture in principle.
For natural language generation, we proposed such an architecture, the Tensor Product Generation Network (TPGN): it embodies the TPR operation of unbinding which is used to extract particular symbols (e.g., words) from complex structures (e.g., sentences).
The architecture can be interpreted as containing a part that encodes a sentence and a part that selects one structural role at a time to extract from the sentence.
We applied the approach to image-caption generation, developing a TPGN model that was evaluated on the COCO dataset, on which it outperformed LSTM baselines on a range of standard metrics.
Unlike standard LSTMs, however, the TPGN model admits a level of interpretability: we can see which roles are being unbound by the unbinding vectors generated internally within the model.
We find such roles contain considerable grammatical information, enabling POS tag prediction for the words they generate and displaying clustering by POS.



\end{document}